\documentclass[letterpaper, 10 pt, conference, twoside]{ieeeconf}  

\IEEEoverridecommandlockouts                              

\usepackage{graphics}
\usepackage[pdftex]{graphicx}
\usepackage[tight,footnotesize]{subfigure}
\usepackage{multirow}
\usepackage[cmex10]{amsmath}
\usepackage{amssymb}
\usepackage{amsfonts}
\usepackage{algpseudocode}
\usepackage{algorithm}
\usepackage{booktabs}
\usepackage{url}
\usepackage{color}
\usepackage{hyperref}
\usepackage{array}
\usepackage{amsmath}
\usepackage{graphicx}
\usepackage{cleveref}
\usepackage{xspace}
\usepackage{cuted}
\usepackage{caption}
\usepackage{etoolbox}
\makeatletter
\patchcmd{\@maketitle}{\end{center}}{\end{center}\vspace*{-0.5cm}}{}{}
\makeatother

\title{\LARGE \bf
GALoc: Gravity Aligned Wireframes for\\
Depth-Free Monocular Floorplan Localization
}

\author{
Jeahn Han$^{1}$, Minji Kim$^{1}$, Jeongbin Sohn$^{1}$, Jonghyeok Park$^{2}$, Matthias Wüest$^{3}$, and Pyojin Kim$^{1}$\\[2pt]
$^{1}$GIST \quad $^{2}$KAIST \quad $^{3}$Zurich University of Applied Sciences
}

\begin{document}

\maketitle

\begin{strip}
  \centering
  \includegraphics[width=0.7\textwidth]{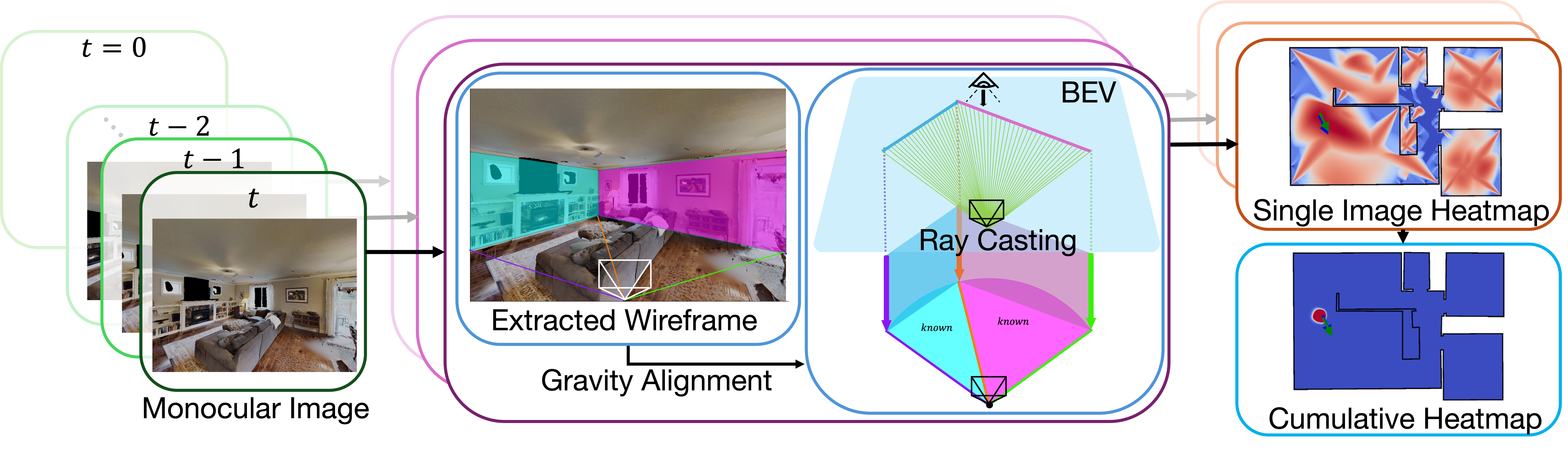}
  \captionof{figure}{Overview of GALoc, multi-view floorplan matching for localizing $\text{SE}(2)$ camera poses with gravity-aligned wireframes.
  GALoc uses gravity-aligned wireframes and wall junctions from RGB images to get bird's-eye-view (BEV) rays.
  GALoc matches them against a floorplan with $\text{SE}(2)$ search and sequential fusion to achieve precise floorplan localization.}
  \label{fig:front_page_figure}
  \vspace{-10pt}
\end{strip}



\noindent \begin{abstract}
Floorplans are compact, appearance-invariant maps ideal for indoor localization, yet existing methods rely on depth networks that are brittle in cluttered scenes. We propose GALoc, a geometry-first framework that replaces depth prediction with gravity-aligned wireframes that satisfy verticality and coplanarity \emph{by construction}. 
Given monocular RGB, camera intrinsics, relative poses, and IMU orientation, GALoc constructs a linear constraint matrix encoding verticality and coplanarity, and finds the camera gauge minimizing its smallest singular value via global search.
The rectified wireframes are projected into bird's-eye-view layouts through a closed-form, FOV-consistent transformation and matched against the floorplan via metric-free SE(2) search.
We evaluate end-to-end on Structured3D, with calibrated noise on Gibson, and on real-world author-collected sequences. When sufficient wall geometry is visible, GALoc matches or outperforms depth-based baselines — achieving 88\% sequential localization success at 0.1m over 100-step sequences on Gibson vs the baseline's 68\% — while abstaining in structure-blind scenes.
\end{abstract}

\section{Introduction}
\label{sec:intro}

\noindent Visual localization underpins 3D reconstruction~\cite{schonberger2016structure}, augmented reality~\cite{sattler2016large}, and autonomous navigation~\cite{boniardi2019robot}, yet classical approaches based on feature matching~\cite{kim2017robust} or depth-derived point clouds~\cite{yin20193d} require dense reconstructions or pre-built maps that degrade under appearance changes.
Floorplan-based localization \cite{workman2015wide,howard2021lalaloc, howard2022lalaloc++, min2022laser} sidesteps these costs: floorplans are compact, appearance-invariant, and encode structural geometry without per-scene retraining.

\noindent However, existing floorplan methods still route through depth---whether from RGB-D sensors~\cite{winterhalter2015accurate} or learned networks~\cite{chen2024f3loc}---matching predicted depth to floorplan structure.
This coupling is fragile: in cluttered environments furniture is mistaken for walls, and in unseen or low-texture scenes depth regressors extrapolate poorly.
The core issue is that depth is an intermediate representation the task does not require; what we need is the \emph{geometric structure of walls}.

\noindent We present GALoc, a depth-free monocular framework that replaces depth prediction with gravity-aligned wireframes (\Cref{fig:front_page_figure}).
Our key insight (\Cref{subsec:ppga}) is that gravity alignment can be cast as an analytical null-space search: we build a linear constraint matrix from 2D wireframe junctions encoding verticality and coplanarity, then recover the unknown camera gauge as the angle minimizing its smallest singular value. 
The recovered 3D wireframe satisfies both constraints exactly by construction, without depth supervision, and is projected into a bird's-eye-view (BEV) layout matched against the floorplan via metric-free SE(2) search.
Our contributions are:
\begin{itemize}
\item[$\bullet$] \textbf{Null Space Gravity-Aligned Rectification}:
a 1D gauge search over yaw selecting the angle minimizing the smallest singular value of the architectural constraint matrix, without depth regression.
\item[$\bullet$] \textbf{Field-of-View Consistent BEV Transformation}:
From rectified wireframes and FOV, we derive a closed-form, scale-free BEV translation using an inscribed-angle constraint that guarantees FOV consistency.
\end{itemize}

\noindent Notably, GALoc requires no learned depth, no semantic annotations, and no panoramic input at the cost of a wireframe extractor (\Cref{tab:input_assumptions}).

\section{Related Works}

\label{sec:related_works}

\noindent\textbf{From Dense Maps to Floorplans.}
Visual localization methods trade off map richness against map cost: image retrieval~\cite{keetha2023anyloc} scales well but depends on database coverage; SfM~\cite{schonberger2016structure,panek2022meshloc} yields 3D maps that must be rebuilt when scenes change; and learning-based methods~\cite{brachmann2017dsac, kendall2015posenet} remain scene-specific. All incur recurring costs, motivating floorplans as an alternative.

\noindent\textbf{Floorplan Localization.}
Floorplan-based methods infer $\text{SE}(2)$ poses by aligning images to architectural maps~\cite{workman2015wide, tian2017cross}. Early approaches match LiDAR or RGB-D depth rays to the floorplan~\cite{yin20193d, zimmerman2022long}, while layout-matching methods extract room edges from RGB~\cite{lin2019floorplan, boniardi2019robot}. RGB-only methods learn image--floorplan compatibility: LaLaLoc/LaLaLoc++~\cite{howard2021lalaloc, howard2022lalaloc++} via shared embeddings, LASER~\cite{min2022laser} via pose-conditioned descriptors, \cite{kim2024fully} via a fully geometric method, and F3Loc~\cite{chen2024f3loc} via learned wall-distance rays---trained to regress floorplan-supervised depth that sees through clutter---fused in a histogram filter. 
Subsequent work enriches this depth-ray model with semantic labels~\cite{grader2025supercharging}, foundation-model depth~\cite{cheng2025palms+}, 3D geometric priors~\cite{chen2025perspective}, or uncertainty~\cite{wuest2025unloc}, but all retain depth as the core representation. We replace depth with wireframe geometry.

\noindent\textbf{Sequential Localization.}
Temporal fusion improves robustness via Bayesian filtering~\cite{dellaert1999monte}: particle filters~\cite{karkus2018particle, boniardi2019robot}, histogram filters~\cite{mendez2020sedar}, and F3Loc~\cite{chen2024f3loc} all fuse depth-derived observations sequentially. We adopt the same histogram-filter backbone but feed it depth-free geometric observations, showing that the observation model is the limiting factor.

\noindent\textbf{Room Layout Estimation.}
The field has progressed from Manhattan-world vanishing points~\cite{ zhang2019edge} through CNN-based prediction~\cite{yan20203d} to direct polygon extraction~\cite{stekovic2020general, gillsjo2023polygon}. We employ PolygonHGT~\cite{gillsjo2023polygon} as our front-end.


\noindent\textbf{Gravity Alignment.} Many systems---including F3Loc~\cite{chen2024f3loc}---apply a homography for gravity alignment, which rectifies the image without imposing verticality or coplanarity on the recovered geometry. 
We instead formulate alignment as a null-space search over depths (\Cref{subsec:ppga}), yielding a 3D structure where both hold by construction, without depth supervision.

\section{Floorplan Localization}

\label{sec:floorplan_localization}

\begin{figure*}[t] 
    \centering
    \includegraphics[width=0.8\textwidth]{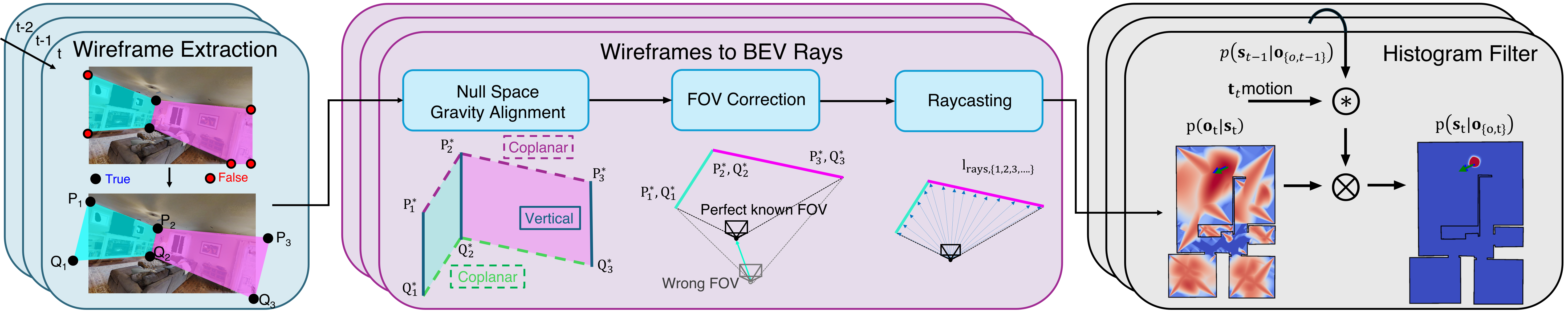} 
    \caption{Our pipeline uses a wireframe extractor (\Cref{subsec:wireframe_extraction}) and corner completion algorithm (\Cref{subsec:corner_completion}) to obtain full walls.
    These walls are gravity-aligned so that verticality and coplanarity hold by construction (\Cref{subsec:ppga}) yielding an accurate BEV representation.
    We then determine the wall–pose relationship based on the FOV (\Cref{subsec:bev}), perform raycasting on the BEV walls and use the resulting rays for floorplan localization (\Cref{subsec:raycasting}).}
    \label{fig:pipeline}
    \vspace{-10pt}
\end{figure*}

\noindent\textbf{Problem Definition.}
Given a temporal sequence of $k{+}1$ monocular RGB images $\mathcal{I} = \{I_\tau \mid \tau \in \{t{-}k,\dots,t\}\}$ with known camera intrinsics $K$, relative poses, and gravity directions (roll/pitch) from an IMU, we seek the $\text{SE}(2)$ camera pose $s_t = [s_{x,t},\, s_{y,t},\, s_{\phi,t}]$ within a wall-only floorplan, represented as a set of 2D line segments encoding walls.

\noindent\textbf{Overview.}
\Cref{fig:pipeline} illustrates our pipeline.
We extract wireframes from each frame (\Cref{subsec:wireframe_extraction}) and complete truncated wall cycles (\Cref{subsec:corner_completion}).
Our core contribution is a null space gravity alignment that recovers gravity-aligned 3D wireframes via null-space search---without depth regression (\Cref{subsec:ppga}).
The rectified wireframe is projected into a bird's-eye-view layout with a closed-form FOV-consistent transformation (\Cref{subsec:bev}), raycast against the floorplan and fused over time in a histogram filter (\Cref{subsec:raycasting}).
\Cref{subsec:ppga} \& \Cref{subsec:bev} are our main contributions.

 \subsection{Wireframe Extraction}
\label{subsec:wireframe_extraction}
\noindent We employ PolygonHGT~\cite{gillsjo2023polygon}, a wireframe extractor, to obtain wall structure from each RGB image.
It returns the pixel coordinates of wall junctions, a connectivity matrix defining edges between them, and ceiling/floor labels for each junction.
We measure wireframe quality via intersection over union, $\mathrm{IoU}(M,\hat{M}) = |M \cap \hat{M}|/|M \cup \hat{M}|$, between the ground-truth layout mask $M$ and the predicted mask $\hat{M}$.

\subsection{From Truncated Cycles to Full Wall Corners}
\label{subsec:corner_completion}

\begin{figure}[t]
    \centering
    \includegraphics[width=0.9\columnwidth]{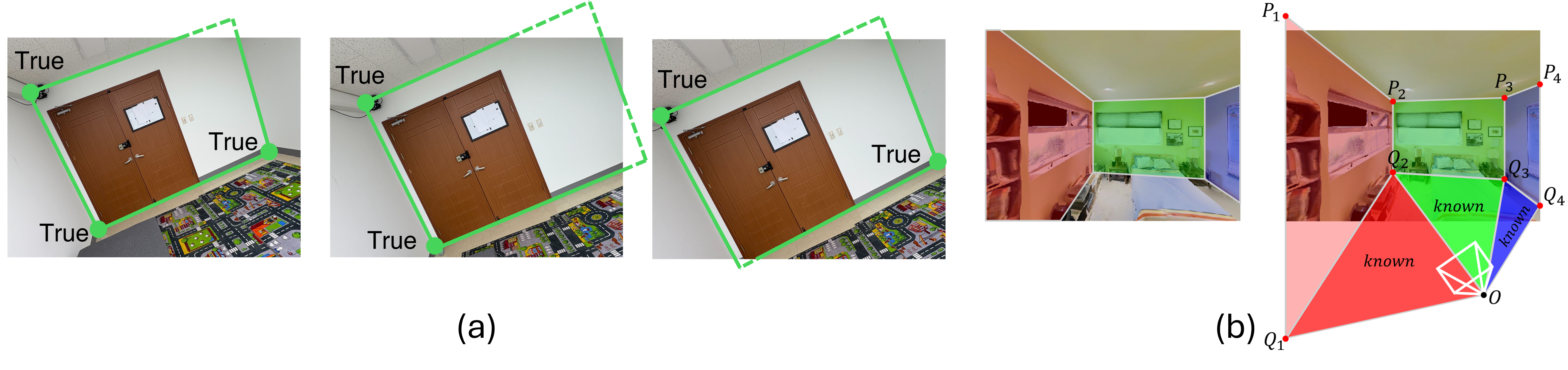}
    \caption{\textbf{(a)} Truncated cycle completion for $j_\mathrm{true}=3$ (left),
    $j_\mathrm{true}=2$ adjacent (middle), and $j_\mathrm{true}=2$ non-adjacent (right).
    \textbf{(b)} After extraction (left), truncated sections are extended (right)
    and FOV regions are labeled as known (K) or unknown (U).}
    \label{fig:truncation_and_fov_labeling}
    \vspace{-10pt}
\end{figure}

\noindent Each visible wall panel is a four-corner cycle. We denote true junctions (actual corners) as $j_\mathrm{true}$.
Corners truncated by the image boundary are completed via line intersections and a vanishing direction for well-posed cases ($j_\mathrm{true}\in\{2,3\}$); under-constrained cases ($j_\mathrm{true}\le 1$) are left unchanged. Each completed cycle receives ceiling ($P$) and floor ($Q$) labels and a horizontal FOV arc label---known (K) if all four corners are recovered, otherwise unknown (U) (\Cref{fig:truncation_and_fov_labeling}).

\subsection{Null Space Gravity Alignment}

\label{subsec:ppga}

\begin{figure}[t]
    \centering
    \includegraphics[width=0.9\columnwidth]{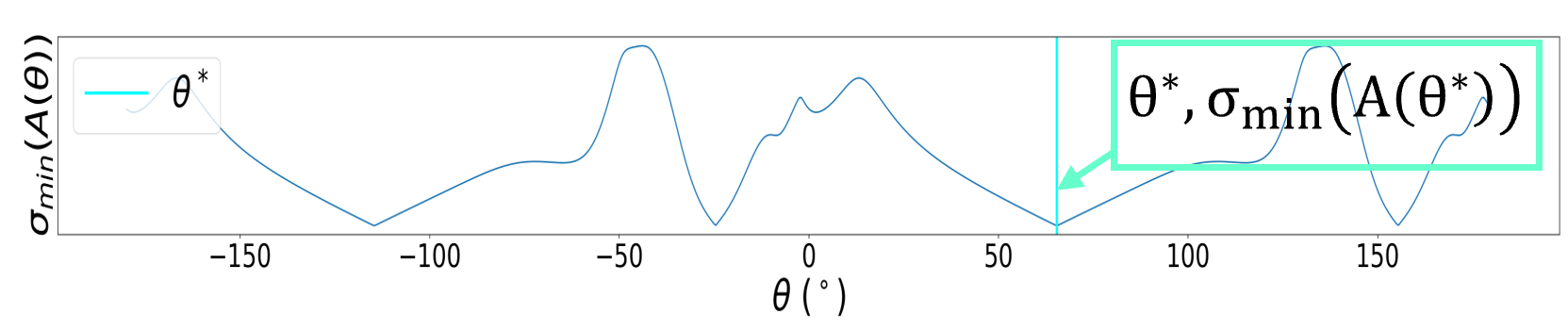}
    \caption{ \textbf{Gauge estimation.} The gauge $\hat\theta$ minimizes $\sigma_{\min}(A(\theta))$, which exhibits four minima at $\frac{\pi}{2}$ intervals.}
    \label{fig:sec3.3_optimization}    
    \vspace{-25pt}
\end{figure}

\noindent\textbf{Formulation.}
We formulate gravity alignment as a null-space search rather than an optimization over reprojection error.
Given the 2D wireframe junctions, we construct a linear constraint matrix $A(\theta)$ encoding verticality and coplanarity (\Cref{tab:constraints}); a valid 3D structure exists only where $A(\theta)$ is rank-deficient.
GALoc proceeds in two stages: Stage~1 estimates the yaw gauge $\hat\theta=\arg\min_{\theta}\sigma_{\min}(A(\theta))$, and Stage~2 fixes $\hat\theta$ and recovers the structure as the unit-norm minimizer $\hat Z=\arg\min_{\|Z\|=1}\|A(\hat\theta)Z\|=\mathbf v_N$, in closed form via Eckart--Young--Mirsky.
The constraints therefore fix the admissible form of the wireframe rather than being solved for, as in a minimal solver \cite{kukelova2008automatic}.


\noindent\textbf{Coordinate Frame.}
We adopt the standard camera convention: $x$~right, $y$~down, $z$~into the scene (optical axis).
Given an IMU-corrected rotation $R_{\mathrm{rp}}$ that removes roll and pitch, the residual unknown is a single yaw angle $\theta$ about gravity.
The full camera-to-world rotation is $R(\theta) = R_y(\theta)R_{\mathrm{rp}}$ where $R_y(\theta)$ is a rotation about the gravity-aligned $y$-axis.
For each ceiling ($P_i$) and floor ($Q_i$) junction with pixel coordinates $\tilde{\mathbf{u}}^P_i$ and $\tilde{\mathbf{u}}^Q_i$, we define gauge-dependent observation rays:

\vspace{-10pt}
\begin{equation}
\small
\label{eq:rays}
\tilde{\mathbf{r}}^P_i(\theta) = R(\theta)\,K^{-1}\tilde{\mathbf{u}}^P_i, \quad
\tilde{\mathbf{r}}^Q_i(\theta) = R(\theta)\,K^{-1}\tilde{\mathbf{u}}^Q_i.
\end{equation}
\vspace{-10pt}

\noindent We suppress the $\theta$-dependence in the notation below for readability.
Every ray $\tilde{\mathbf{r}} = (\tilde r_x,\, \tilde r_y,\, \tilde r_z)^\top$ follows the $(x,y,z)$ convention defined above.

\noindent\textbf{Depth Parameterization.}
Each ceiling point $P_i = [x_i^P,\, y_i^P,\, z_i^P]^\top$ lies along its observation ray at an unknown scalar depth:
\begin{equation}
\small
\label{eq:P}
{P}_i = Z_{P_i}\,\tilde{\mathbf{r}}^P_i,
\end{equation}
where $Z_{P_i} > 0$ is the depth (scale along the ray) in the camera frame. 
Recall that each ray $\tilde{\mathbf{r}}^B_i(\theta) = R(\theta)\,K^{-1}[u_i^B,\,v_i^B,\,1]^\top$ from \Cref{eq:rays}. The $k$ coordinate ($k \in \{x,y,z\}$) of a 3D point $B_i$ ($B \in \{P,Q\}$) is then
\begin{equation}
\small
\label{eq:component}
k_i^B \;=\; \tilde r^B_{i,k}\, Z_{B_i},
\end{equation}
where $\tilde r^B_{i,k}$ denotes the $k$-th component of $\tilde{\mathbf{r}}^B_i$.
\noindent We now eliminate the floor depth $Z_{Q_i}$ as an independent variable.
A vertical wall requires $P_i$ and $Q_i$ to share the same $z$-coordinate:

\vspace{-5pt}
\begin{equation}
\scriptsize
\label{eq:vert_z}
\begin{split}
z_i^P = z_i^Q
\quad\Rightarrow\quad
Z_{P_i}\,\tilde r^P_{i,z} = Z_{Q_i}\,\tilde r^Q_{i,z}\\
\quad\Rightarrow\quad
Z_{Q_i} = \frac{\tilde r^P_{i,z}}{\tilde r^Q_{i,z}}\,Z_{P_i} 
= \lambda_i\,Z_{P_i}, \;\;
\lambda_i \;\triangleq\; \frac{\tilde r^P_{i,z}}{\tilde r^Q_{i,z}}.
\end{split}
\end{equation}

\noindent Substituting back, the floor point is

\vspace{-5pt} 
\begin{equation}
\small
\label{eq:Q}
{Q}_i = \lambda_i\,Z_{P_i}\,\tilde{\mathbf{r}}^Q_i.
\end{equation}
The structure recovery problem reduces to finding the vector $Z = [Z_{P_1},\dots,Z_{P_N}]^\top$, where $N$ is the number of ceiling--floor vertical pairs as the variable $Z_{Q_i}=\lambda_i\,Z_{P_i}$ is now a function of $Z_{P_i}$.

\noindent\textbf{Architectural Constraints.}
We derive three families of linear constraints on $Z$ from indoor geometry. These constraints are shown in \Cref{tab:constraints}.

\noindent\emph{(i) Verticality ($x$-coordinate).}
A vertical wall requires the ceiling and floor points to also share the same $x$-coordinate:
\begin{equation}
\scriptsize
\label{eq:vert_x}
x_i^P = x_i^Q
\;\;\Rightarrow\;\;
Z_{P_i}\,\tilde r^P_{i,x} = \lambda_i\,Z_{P_i}\,\tilde r^Q_{i,x}
\;\;\Rightarrow\;\;
\bigl(\tilde r^P_{i,x} - \lambda_i\,\tilde r^Q_{i,x}\bigr)\,Z_{P_i} = 0.
\end{equation}

\noindent This row involves the single unknown $Z_{P_i}$, the floor depth having been eliminated via $Z_{Q_i}=\lambda_i Z_{P_i}$ (\Cref{eq:vert_z}): the ceiling and floor rays enter through the single coefficient $(\tilde{r}^P_{i,x}-\lambda_i\,\tilde{r}^Q_{i,x})$ rather than through separate columns of $A(\theta)$. 
This coefficient depends on $\theta$ but vanishes at the correct gauge regardless of $Z_{P_i}$, so this row constrains $\theta$ rather than the structure.
The rows are nonetheless retained for two reasons: they sharpen the $\sigma_{\min}$ basin, aiding Brent convergence; and under noise ($\hat\theta\neq\theta^\ast$) the coefficient is small but nonzero. 
So they enter the EYM projection and shape the recovered depths jointly with the coplanarity constraints. 
Without them, nothing in $A(\theta)$ couples the $x$-coordinates of matched ceiling/floor points, and $Z$ is fit by coplanarity alone.


\noindent\emph{(ii) Ceiling coplanarity ($y$-coordinate).}
All ceiling points lie on a horizontal plane, so adjacent ceiling points $P_i$ and $P_{i+1}$ share the same height:

\vspace{-5pt}
\begin{equation}
\scriptsize
\label{eq:ceil_coplanar}
\begin{split}
y_i^P = y_{i+1}^P
\;\;\Rightarrow\;\;
Z_{P_i}\,\tilde r^P_{i,y} = Z_{P_{i+1}}\,\tilde r^P_{i+1,y} \\
\;\;\Rightarrow\;\;
\tilde r^P_{i,y}\,Z_{P_i} - \tilde r^P_{i+1,y}\,Z_{P_{i+1}} = 0.
\end{split}
\end{equation}

\noindent\emph{(iii) Floor coplanarity ($y$-coordinate).}
Similarly, all floor points share a common height.
Substituting $Z_{Q_i} = \lambda_i\,Z_{P_i}$ from \Cref{eq:vert_z}:

\vspace{-5pt}
\begin{equation}
\scriptsize
\label{eq:floor_coplanar}
\begin{split}
y_i^Q = y_{i+1}^Q
\;\;\Rightarrow\;\; 
\lambda_i\,Z_{P_i}\,\tilde r^Q_{i,y} = \lambda_{i+1}\,Z_{P_{i+1}}\,\tilde r^Q_{i+1,y} \\
\;\;\Rightarrow\;\;
\tilde r^Q_{i,y}\,\lambda_i\,Z_{P_i} - \tilde r^Q_{i+1,y}\,\lambda_{i+1}\,Z_{P_{i+1}} = 0.
\end{split}
\end{equation}

\noindent For $N$ vertical wall pairs, $N$ verticality rows, $N{-}1$ ceiling coplanarity rows, and $N{-}1$ floor coplanarity rows in \Cref{tab:constraints}, form the $(3N{-}2)\times N$ matrix $A(\theta)$.

\noindent BEV projection (\Cref{subsec:bev}) requires the full depth vector $Z$, not just the gauge $\theta$; we therefore retain $Z$ as an explicit variable, so a single SVD of $A(\theta)$ couples gauge estimation and structure recovery.
Eliminating $Z$ via cross-multiplication is less practical: $\sigma_{\min}(A(\theta))$ is defined implicitly as the smallest root of $\det(A^{\top} A - \lambda I) = 0$, which is non-polynomial in $\theta$, and algebraic elimination via resultants yields a univariate polynomial of degree $O(N^2)$ offering no practical advantage over our 360-point scan below.


\begin{table}[t]
\scriptsize
\caption{Architectural constraints forming $A(\theta)Z=0$.
Here $Z=[Z_{P_1},\dots,Z_{P_N}]^\top$ are depths in the camera frame,
$\lambda_i = \tilde r^P_{i,z}/\tilde r^Q_{i,z}$ couples each floor depth to its ceiling depth (\Cref{eq:vert_z}),
and all rays $\tilde{\mathbf{r}}^{P,Q}_i(\theta)$ depend on the unknown yaw $\theta$ via \Cref{eq:rays}.}
\centering
\renewcommand{\arraystretch}{1.2}
\setlength{\tabcolsep}{3pt}
\begin{tabular}{@{}l l l@{}}
\textbf{Constraint} & \textbf{Geometry} & \textbf{Linear row in $Z$} \\
\hline
Verticality &
$x_i^P=x_i^Q$ &
$(\tilde r^P_{i,x} - \lambda_i\,\tilde r^Q_{i,x})\,Z_{P_i} = 0$ \\[3pt]
Ceiling coplanarity &
$y_i^P=y_{i+1}^P$ &
$\tilde r^P_{i,y}\,Z_{P_i} - \tilde r^P_{i+1,y}\,Z_{P_{i+1}} = 0$ \\[3pt]
Floor coplanarity &
$y_i^Q=y_{i+1}^Q$ &
$\tilde r^Q_{i,y}\,\lambda_i\,Z_{P_i} - \tilde r^Q_{i+1,y}\,\lambda_{i+1}\,Z_{P_{i+1}} = 0$
\end{tabular}
\label{tab:constraints}
\vspace{-15pt}
\end{table}

\noindent\textbf{Gauge Estimation.}
At the correct yaw $\theta^*$, the architectural constraints are simultaneously satisfiable with ideal pixel coordinates: a non-trivial structure $Z\neq 0$ exists such that $A(\theta^*)Z=0$.
Equivalent to $A(\theta^*)$ being rank-deficient:

\begin{equation}
\small
\label{eq:rank_condition}
\exists\,Z\neq 0:\;A(\theta)Z=0
\;\;\iff\;\;
\sigma_{\min}\!\bigl(A(\theta)\bigr)=0.
\end{equation}

\noindent When noise-free, $\sigma_{\min}(A(\theta^*))=0$.
Under pixel noise, the rays $\tilde{\mathbf{r}}^{P,Q}_i$ are perturbed and $A(\theta)$ becomes generically full-rank for all $\theta$; no angle makes the constraints exactly satisfiable.
We recover the best gauge by finding the angle that minimizes the smallest singular value ($\sigma_{min}(A(\theta))$): 

\begin{equation}
\label{eq:theta_star}
\small
\hat\theta \;=\; \arg\min_{\theta\in[-\pi,\pi]}\;\sigma_{\min}\!\bigl(A(\theta)\bigr).
\end{equation}

\noindent The landscape $\sigma_{\min}(A(\theta))$ has four minima at $\tfrac{\pi}{2}$ spacing. 
Since $R_y(\theta)$ fixes the $y$-component of every ray, $\theta$ enters $A(\theta)$ only through the verticality rows and through $\lambda_i$ in the floor-coplanarity rows. 
A yaw of $\pi$ maps $\tilde{\mathbf r}\mapsto(-\tilde r_x,\tilde r_y,-\tilde r_z)$: $\lambda_i$ is invariant, both coplanarity families are unchanged, and the verticality rows negate, so $A(\theta+\pi)=D\,A(\theta)$ with $D=\mathrm{diag}(\pm1)$ orthogonal. 
The spectrum is therefore \emph{exactly} $\pi$-periodic, for every $\theta$ and every noise level, and a $180^\circ$-rotated room is algebraically indistinguishable.
A yaw of $\tfrac{\pi}{2}$ maps $\tilde{\mathbf r}\mapsto(\tilde r_z,\tilde r_y,-\tilde r_x)$, so $\lambda_i(\theta+\tfrac{\pi}{2})=\tilde r^{P}_{i,x}/\tilde r^{Q}_{i,x}$. 
At the true gauge a vertical wall satisfies $x^P_i=x^Q_i$ as well as $z^P_i=z^Q_i$, forcing $\tilde r^{P}_{i,x}/\tilde r^{Q}_{i,x}=Z_{Q_i}/Z_{P_i}=\lambda_i(\theta^\ast)$: the coplanarity rows are again unchanged while the verticality coefficient
$\tilde r^{P}_{i,z}-\lambda_i\tilde r^{Q}_{i,z}$ vanishes. 
Hence $A(\theta^\ast+\tfrac{\pi}{2})=A(\theta^\ast)$ in the noise-free case---the solver cannot distinguish facing a wall head-on from looking along it.
The $\tfrac{\pi}{2}$ structure makes the search global: a $1^\circ$ scan of $\sigma_{\min}$ over $[-\pi,\pi]$ brackets four minima, and Brent refinement~\cite{brent2013algorithms} within a $\pm 1^\circ$ bracket recovers sub-degree precision.



\noindent\textbf{Structure Recovery Using Null Space Gravity Alignment.}
At $\hat\theta$, the matrix $A(\hat\theta)$ is full-rank due to noise---the architectural constraints are only \emph{approximately} satisfied.
We enforce \emph{exact} constraint satisfaction via the Eckart--Young--Mirsky (EYM) theorem~\cite{golub1987generalization}.
Let $A(\hat\theta) = U\Sigma V^\top$ be the SVD.
We construct the nearest rank-deficient matrix $A'(\hat\theta)$ by zeroing the smallest singular value:
\vspace{-15pt}

\begin{equation}
\scriptsize
\label{eq:eym}
\setlength{\arraycolsep}{2pt}
A'(\hat\theta) = U\Sigma' V^\top, \;\;
\Sigma = \begin{pmatrix}
\sigma_1 & \dots & 0 \\
\vdots & \ddots & \vdots \\
0 & \dots & \sigma_{\min}
\end{pmatrix}
\!\rightarrow\!
\Sigma' = \begin{pmatrix}
\sigma_1 & \dots & 0 \\
\vdots & \ddots & \vdots \\
0 & \dots & 0
\end{pmatrix}
\end{equation}



\noindent where $\Sigma'$ replaces $\sigma_{\min}$ with zero.
\Cref{eq:eym} shows the relationship between $\Sigma$ and $\Sigma'$.
By construction, $A'(\hat\theta)$ is rank-deficient and its null-space vector

\vspace{-10pt}

\begin{equation}
\small
\label{eq:Z_recovery}
Z(\hat\theta) = \mathbf{v}_N, \quad \text{where } \mathbf{v}_N \text{ is the last column of } V,
\end{equation}

\noindent satisfies $A'(\hat\theta)Z = 0$ exactly.
Verticality and coplanarity are imposed by the depth parameterization and the assembly below, so they hold for \emph{any} $Z$: the constraints fix the admissible form of the wireframe rather than being solved for, as in a minimal solver \cite{kukelova2008automatic}. 
What the EYM step supplies is the depths. 
The null space of the Frobenius-nearest rank-deficient $A'(\hat\theta)$ gives the $Z$ most consistent with the measured rays. 
The exactness is therefore a property of the recovered 3D structure, not of the input pixels.

\noindent\textbf{Scope of the guarantee.}
The EYM step minimizes $\|A(\hat\theta) - A'(\hat\theta)\|_F$, a Frobenius-norm perturbation of the \emph{constraint matrix}, not a geometric reprojection error in pixel space.
We therefore do not claim minimality with respect to pixel or ray coordinates: the Frobenius-optimal $A'(\hat\theta)$ need not correspond to any realizable adjustment of the rays, since a given ray component appears in several rows that EYM perturbs jointly.
What $A'(\hat\theta)$ provides is the nearest rank-deficient constraint matrix (Frobenius norm), whose null space yields depths under which the architectural constraints are exactly satisfiable on the recovered structure.


\noindent\textbf{Coordinate assembly.}
The null-space solve yields depth ratios $Z = \mathbf{v}_N$ (normalized so $Z_1 = 1$); the final coordinates are then assembled so the architectural constraints hold exactly. For each pair $i$, the horizontal coordinates are taken from the ceiling ray $\tilde r^{P}_i(\hat\theta)$ scaled by its depth ratio, and shared with the floor point to enforce verticality:

\begin{equation}
\small
\label{eq:xz_assembly}
x_i = z_1 Z_i\,\tilde r^{P}_{i,x}(\hat\theta), \qquad
z_i = z_1 Z_i\,\tilde r^{P}_{i,z}(\hat\theta),
\end{equation}

\noindent 
where $z_1$ is an optional global scale (set to $1$ due to no metric value). 
The vertical coordinates are set to a single constant per plane, recovered from the coplanarity rows of $A'(\hat{\theta})$,

\begin{equation}
\small
\label{eq:y_assembly}
y_i^P = z_1 Z_1 \, \tilde{r}^P_{1,y}, \qquad
y_i^Q = z_1 Z_1 \, \lambda_1 \, \tilde{r}^Q_{1,y},
\end{equation}

\noindent so all floor (resp.\ ceiling) points are exactly coplanar. 
Thus the solve provides depth ratios, while the assembly enforces verticality (shared $x, z$) and coplanarity (constant $y$) by construction; the input pixels are unchanged. 
Here the plane heights are set from the leading coefficients of the first ceiling-coplanarity and first floor-coplanarity rows of $A(\hat{\theta})$ (Table~\ref{tab:constraints}): the ceiling height from $\tilde{r}^P_{1,y}$, and the floor height from $\lambda_1 \tilde{r}^Q_{1,y}$, where $\lambda_1$ carries the ceiling-to-floor depth coupling of Equation~\eqref{eq:vert_z}. 
Scaled by $Z_1$, each gives the shared plane height its row enforces, so every $P$ (resp.\ $Q$) point inherits the same $y$.

\noindent\textbf{Resolving Gauge Symmetries.}
$\sigma_{\min}(A(\theta))$ has four minima at $\frac{\pi}{2}$ spacing, yielding four candidate angles $\Theta_4$. 
We disambiguate via two physical checks on the assembled structure, not on $Z$ itself: since $A(\theta+\pi)=D\,A(\theta)$ with $D$ orthogonal, $Z(\theta+\pi)=Z(\theta)$, and the $\pi$ pair is separated only by the sign of $\tilde r_z$.

\begin{enumerate}[]
    \small
    \item \textbf{Positive depth:} select $\theta\in\Theta_4$ minimizing
    $\#\{i: z_i(\theta)<0\}$, with $z_i=z_1 Z_i\,\tilde r^{P}_{i,z}(\theta)$.
    \item \textbf{Chirality:} if ties remain, select $\theta$ minimizing lateral
    ordering inconsistencies between image-space and reconstructed 3D points.
\end{enumerate}

\begin{figure}[H]
    \vspace{-10pt}
    \centering
    \includegraphics[width=\columnwidth]{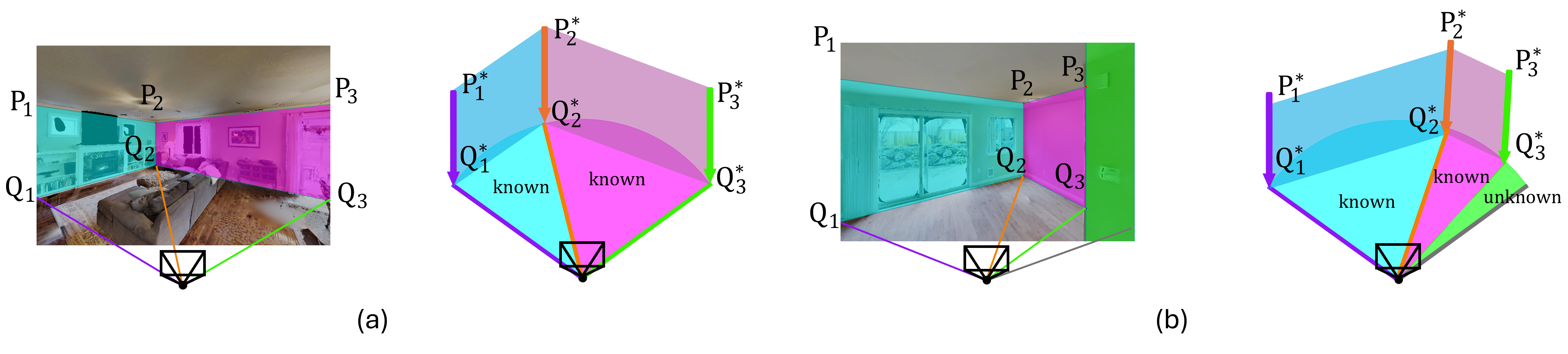}
    \caption{
    \textbf{(a)} Where all walls in the FOV have four true junctions.
    \textbf{(b)} Where \emph{not} all walls have four true corners.}
    \label{fig:ppga_diagram}
    \vspace{-10pt}
\end{figure}

\noindent The resulting pair $(\hat\theta, Z)$ yields a gravity-aligned 3D wireframe that enforces verticality and coplanarity by construction, used for downstream BEV projection and localization (solvable and partial-FOV cases are illustrated in \Cref{fig:ppga_diagram}).


 \subsection{Bird's Eye View (BEV)}
\label{subsec:bev}



\begin{figure}[H]
    \centering
    \includegraphics[width=0.9\columnwidth, trim=10pt 10pt 15pt 5pt, clip]{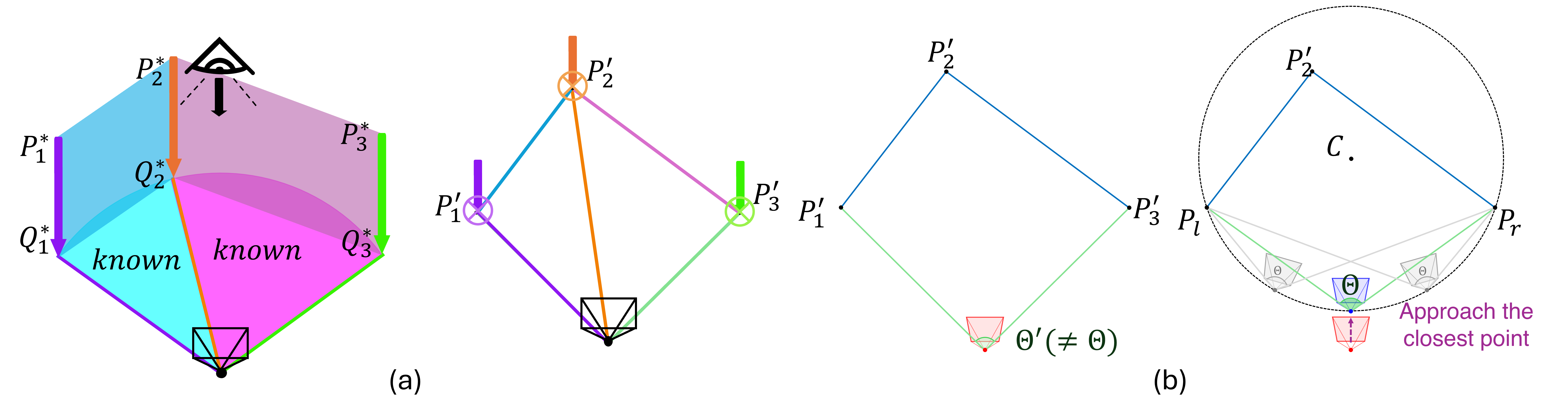}
    \caption{\textbf{Bird's-eye-view construction.}
    \textbf{(a)}~Projecting the gravity-aligned wireframe onto the $X$--$Z$ plane yields a scale-free BEV wall chain. \textbf{(b)}~Under noise the raw BEV camera subtends an incorrect angle $\Theta'\!\neq\!\Theta$ (left); the inscribed-angle theorem places it on a circle where $\mathrm{FOV}=\Theta$ holds exactly (right), and we select the arc point nearest the origin.}
    \label{fig:3d_points_BEV_inscribed}
    \vspace{-10pt}
\end{figure}

\noindent The gravity-aligned 3D wireframe from \Cref{subsec:ppga} is projected onto the $X$--$Z$ plane, giving a BEV wall chain $\{P_i'\}_{i=1}^N$ with the camera at the origin.
Let $P_\ell = P_1'$ and $P_r = P_N'$ be the outermost known-arc endpoints, $d = \|P_r - P_\ell\|$ their chord length, and $\Theta$ the horizontal angle the known arc subtends in the image (equal to the camera FOV when every panel is known).
Under noise, the angle $\Theta'$ that the BEV chain subtends at the origin differs from $\Theta$, distorting ray scoring.
By the inscribed-angle theorem, every point subtending $\Theta$ on $\overline{P_\ell P_r}$ lies on a circle of radius $R = d/(2\sin\Theta)$ whose center $C$ lies on the perpendicular bisector of the chord; we retain the circle on which the camera faces the wall chain and place the camera position at the point of the arc nearest the origin, which is given by $X^\star(C) = C - R(C / \|C\|)$ where $C$ is the center of the retained circle.
This adjusts only the translation, leaving $\hat\theta, Z$ unchanged, and is scale-free: $R \propto d$ inherits the global scale ambiguity of $Z(\hat\theta)$, which cancels in the cosine-similarity scoring of \Cref{subsec:raycasting}.
Only known arcs contribute observation rays; unknown FOV segments are appended at their angular positions but excluded from $\mathcal{K}$ (\Cref{fig:3d_points_BEV_inscribed}).


 \subsection{Raycasting and Floorplan Matching}
\label{subsec:raycasting}

\begin{figure}[t]
    \centering
    \includegraphics[width=0.8\columnwidth]{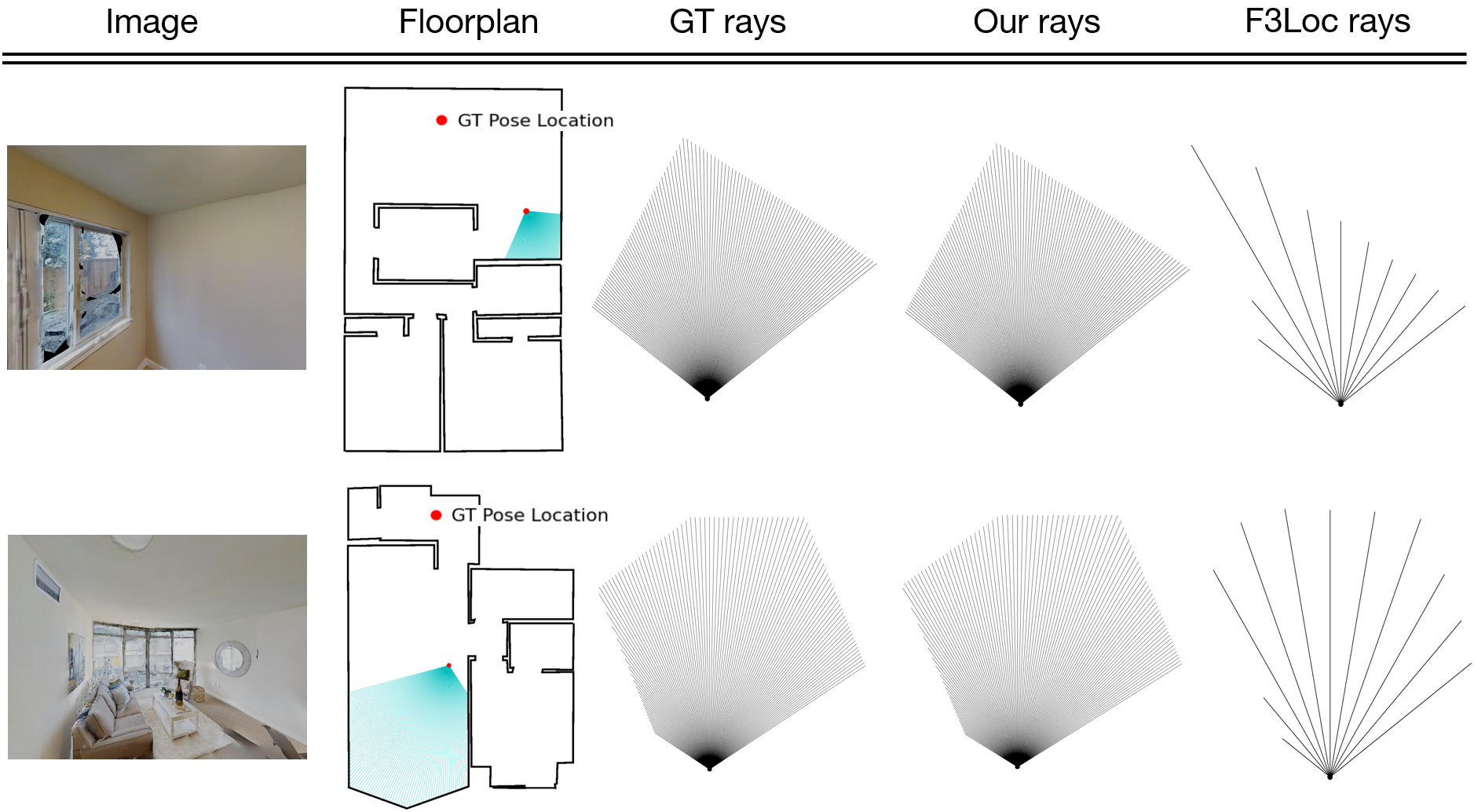}
    \caption{\textbf{Raycast comparison.} GALoc produces ray patterns that closely
    match the ground-truth rays; F3Loc fails to recover accurate wall
    boundaries in these scenarios.}
    \label{fig:ray_comparison}
    \vspace{-20pt}
\end{figure}

\noindent The gravity alignment gauge $\theta$ from \Cref{subsec:ppga} rectifies the wireframe into a canonical frame; the floorplan heading $s_{\phi,t}$ is a separate unknown recovered by the SE(2) search below.
From the BEV wall map and known FOV sectors we cast observation rays and score them against precomputed map profiles to produce a per-pose likelihood. 
GALoc's raycast observations closely track the ground-truth rays, whereas F3Loc misplaces wall boundaries, shown in \Cref{fig:ray_comparison}.

\noindent\textbf{Observation Rays.}
From gravity-aligned junctions we cast $N_r = \lfloor \Theta \rfloor + 1$ rays $l_{\text{rays},i}$ at $1^\circ$ spacing across the horizontal FOV $\Theta$, symmetric about the optical axis, keeping those in known sectors, indexed by $\mathcal{K} \subset \{0,\ldots,N_r-1\}$.
For each grid pose $p=(x,y)$ and heading $v$, we align each ray to its bin in the precomputed 360-bin profile $w_{p,v}$ via $a(i,v) \equiv v + i - \tfrac{N_r-1}{2} \pmod{360}$, yielding the aligned pairs

\begin{equation}
\small
\mathcal{R}(p,v) \;=\; \bigl\{\, (\,l_{\text{rays},i},\, w_{p,v}[a(i,v)]\,) \;:\; i\in \mathcal{K} \bigr\}.
\end{equation}
Figure~\ref{fig:ray_comparison} illustrates the agreement between observed and map-based rays.

\noindent\textbf{Known-Only Cosine Similarity.}
We score each pose with a masked cosine similarity over the known rays, sharpened by $\gamma = 10$:
\begin{equation}
\scriptsize
\label{eq:masked-cos}
s(p,v) = \left[ \max\!\left( 0,\; \frac{\sum_{i \in K} l_{\text{rays},i} \cdot w_{p,v}[a(i,v)]}
{\bigl(\sum_{i\in K} l_{\text{rays},i}^2\bigr)^{\!\frac{1}{2}}\,
 \bigl(\sum_{i\in K} w_{p,v}[a(i,v)]^2\bigr)^{\!\frac{1}{2}} + 10^{-6}} \right) \right]^{\!\gamma}\!.
\end{equation}
Observations with unknown arcs are downweighted by a confidence factor $c = (|\mathcal{K}|/N_r)^2$, floored at $c_{\min}=0.01$.

\noindent\textbf{Sequential Fusion.}
We maintain a log-domain belief $\log B_t(p,v)$ over the SE(2) grid.
At each timestep, a prediction step shifts and rolls the previous belief by the odometry $(\Delta x_t, \Delta y_t, \Delta v_t)$, and a measurement step adds the confidence-weighted log-likelihood:

\vspace{-10pt}
\begin{equation}
\small
\log B_t(p,v) \;\leftarrow\; \log \tilde{B}_t(p,v) \;+\; c\,\log\bigl(\varepsilon + s_t(p,v)\bigr),
\end{equation}

\noindent where $\log \tilde{B}_t$ is the motion-propagated prior and $\varepsilon = 0.01$ is a small constant that keeps the logarithm finite when $s_t(p,v) = 0$.
For visualization we marginalize over heading to obtain a 2D pose heatmap, $H_t(p) \propto \sum_v \exp(\log B_t(p,v))$.

\section{Experiments and Results}

\label{sec:experiments}

\noindent We evaluate GALoc in three complementary settings: end-to-end with PolygonHGT on Structured3D, on real-world custom datasets which validate the full pipeline, and with calibrated noise on Gibson.

\noindent\textbf{Evaluation protocol.}
For single-image localization we report \emph{recall} (percentage of poses within $X$\,m of GT) and a new metric \emph{rank} (GT percentile in the output heatmap). 
For sequential localization we report \emph{success rate (SR)}: the pose must stay within $X$\,m of GT over 10 consecutive frames.
Rank remains informative under geometric ambiguities where multiple poses are equally plausible.

\noindent\textbf{Structured3D~\cite{zheng2020structured3d}.}
A synthetic dataset with non-upright viewpoints, low--medium occlusion, and $\sim\!80^{\circ}$ horizontal FOV.
All Structured3D results use PolygonHGT~\cite{gillsjo2023polygon} predictions end-to-end.

\noindent\textbf{Gibson~\cite{xia2018gibson}.}
A synthetic dataset with gravity-aligned cameras and $\sim\!108^{\circ}$ horizontal FOV.
Its floorplans are rasters with wall artifacts that make off-the-shelf corner detectors misfire, so we mark wall junctions manually and use the ground-truth pose to select the subset visible and unoccluded from each viewpoint, projecting them into the frame with a fixed wall and camera height. 
The ground-truth pose determines only \emph{which} junctions a detector would see; it never enters $A(\theta)$, the BEV projection, or the pose scoring, and the heights are used only to build the input wireframe.
To simulate detector imperfection we perturb each ceiling and floor junction \emph{independently} in the image plane with zero-mean Gaussian noise ($\sigma \in \{3,5,7\}$\,px), so verticality and coplanarity no longer hold in the input rays and $A(\theta)$ is full-rank at every $\theta$.
The perturbed junctions yield layout-mask IoUs of 96.76\%, 94.73\%, and 92.78\%; we adopt $\sigma{=}5$, whose 94.73\% matches PolygonHGT's 94.66\% on Structured3D.
Running PolygonHGT end-to-end here would instead conflate geometry with detector domain transfer: Gibson provides no ground-truth wireframes, so PolygonHGT cannot be trained on it, while F3Loc's released weights were.
The comparison is thus conservative for GALoc on the domain axis and favorable on the input-quality axis, since our junctions have correct topology, connectivity, and ceiling/floor labels by construction.


\begin{table}[t]
\scriptsize
\centering
\caption{Method input assumptions. GALoc uses no learned depth, semantics, or panoramas, at the cost of requiring a wireframe extractor.}
\label{tab:input_assumptions}
\setlength{\tabcolsep}{5pt}
\begin{tabular}{lcccc}
 & F3Loc & SemRayLoc & PALMS+ & GALoc \\
\midrule
Depth source & Learned rays & Learned rays & Foundation depth & None \\
Wireframe  & None & None & None & PolygonHGT \\
Floorplan type & Geometric & Semantic & Geometric & Geometric \\
\end{tabular}
\vspace{-5pt}
\end{table}

\noindent\textbf{Baselines.}
We compare GALoc against three baselines: F3Loc~\cite{chen2024f3loc} (learned wall-distance rays, histogram filter), SemRayLoc~\cite{grader2025supercharging} (learned semantic rays, semantic floorplans), and PALMS+~\cite{cheng2025palms+} (foundation-model depth, point-cloud matching). GALoc uses none of these: no learned depth, no semantic annotations, no panoramic input. \Cref{tab:input_assumptions} summarizes each method's input requirements.

\subsection{Single Image Localization}
\label{subsec:single_image_loc}

\begin{figure}[t]
    \centering
    \includegraphics[width=0.8\columnwidth]{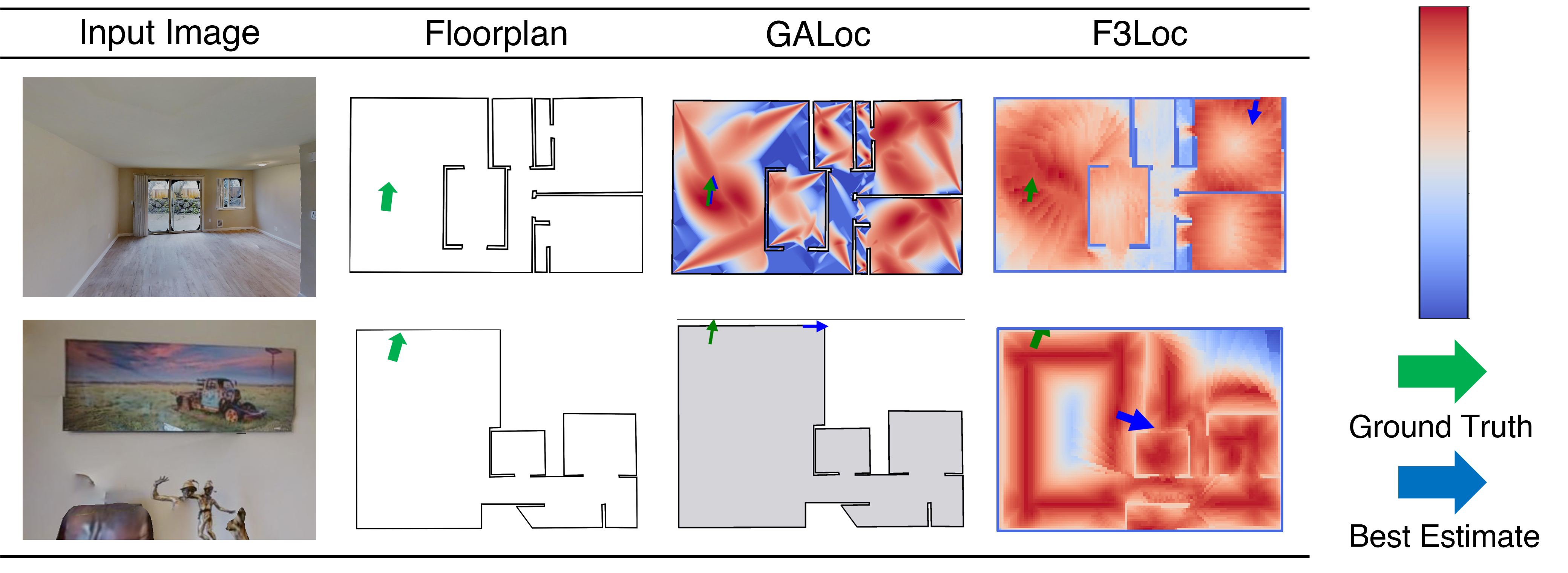}
    \caption{\textbf{Single-image localization.}
    Top: visible wall geometry enables accurate localization.
    Bottom: a featureless close-up yields zero confidence; the near-uniform heatmap injects no information, and the filter falls back to odometry.}
    \label{fig:single_image_localization}
    \vspace{-20pt}
\end{figure}

\begin{table*}[htbp]
  \scriptsize
  \centering
  \caption{Single-image recall and rank on Structured3D~\cite{zheng2020structured3d} and Gibson~\cite{xia2018gibson}. GALoc's lower aggregate recall on Structured3D reflects the ${\sim}40\%$ of frames with zero wall coverage (\Cref{fig:coverage_combined}); on Gibson, where over 80\% of frames have sufficient coverage, GALoc achieves $3{\times}$ the recall of $\text{F3Loc}_m$ at 0.1\,m.}
  \label{tab:combined_localization_unified}
  \setlength{\tabcolsep}{2.2pt} 
  \resizebox{\textwidth}{!}{%
  \begin{tabular}{l cccccc cccc @{\hspace{16pt}} l cccccc cccc}
    & \multicolumn{10}{c}{\textbf{Structured3D~\cite{zheng2020structured3d}}} & & \multicolumn{10}{c}{\textbf{Gibson~\cite{xia2018gibson}}} \\
    \cmidrule(lr){2-11} \cmidrule(lr){13-22}
    & \multicolumn{6}{c}{\textbf{Recall (Thresholds)}} & \multicolumn{4}{c}{\textbf{Rank (\%)}} & & \multicolumn{6}{c}{\textbf{Recall (Thresholds)}} & \multicolumn{4}{c}{\textbf{Rank (\%)}} \\
    \cmidrule(lr){2-7} \cmidrule(lr){8-11} \cmidrule(lr){13-18} \cmidrule(lr){19-22}
    \textbf{Method} & 0.1m & 0.3m & 0.5m & 0.7m & 1m, $30^\circ$ & 1m & 0.01 & 0.02 & 0.05 & 0.10 & \textbf{Method} & 0.1m & 0.3m & 0.5m & 0.7m & 1m, $30^\circ$ & 1m & 0.01 & 0.02 & 0.05 & 0.10 \\
    \midrule
    $\text{F3Loc}_{m}$ \cite{chen2024f3loc} & 1.2 & 8.7 & 12.1 & 18.5 & 20.0 & 21.9 & 2.7 & 13.0 & 19.8 & 23.0 & $\text{F3Loc}_{m}$ & 6.1 & 26.5 & 37.0 & 43.5 & 46.2 & 48.1 & 19.8 & 26.4 & 36.5 & 45.7 \\
    $\text{PALMS}+$ \cite{cheng2025palms+}    & 0.8 & 3.8 & 5.8  & 8.4  & 10.3 & 12.6  & 0.4 & 0.9  & 1.3  & 1.6  & $\text{GALoc}_{(\sigma=3)}$ & 19.4 & 46.8 & 49.6 & 51.4 & 52.0 & 52.8 & 47.6 & 56.1 & 67.0 & 70.2 \\
    $\text{SemRayLoc}$ \cite{grader2025supercharging} & 1.0 & 10.2& 13.1 & 20.2 & 21.8 & 23.8 & 2.3 & 9.5  & 14.0 & 16.1 & $\text{GALoc}_{(\sigma=5)}$ & 18.1 & 44.4 & 48.3 & 50.1 & 50.5 & 51.5 & 45.3 & 52.6 & 63.5 & 69.2 \\
    $\text{GALoc}$ (Ours)                      & 0.3 & 5.1 & 7.3  & 11.9 & 15.7 & 18.2 & 0.3 & 9.2  & 15.8 & 19.4 & $\text{GALoc}_{(\sigma=7)}$ & 14.2 & 39.8 & 45.0 & 46.9 & 47.8 & 48.7 & 43.4 & 51.5 & 61.7 & 67.4 \\
  \end{tabular}%
  }
\end{table*}

\begin{figure}[t]
    \centering
    \includegraphics[width=\columnwidth]{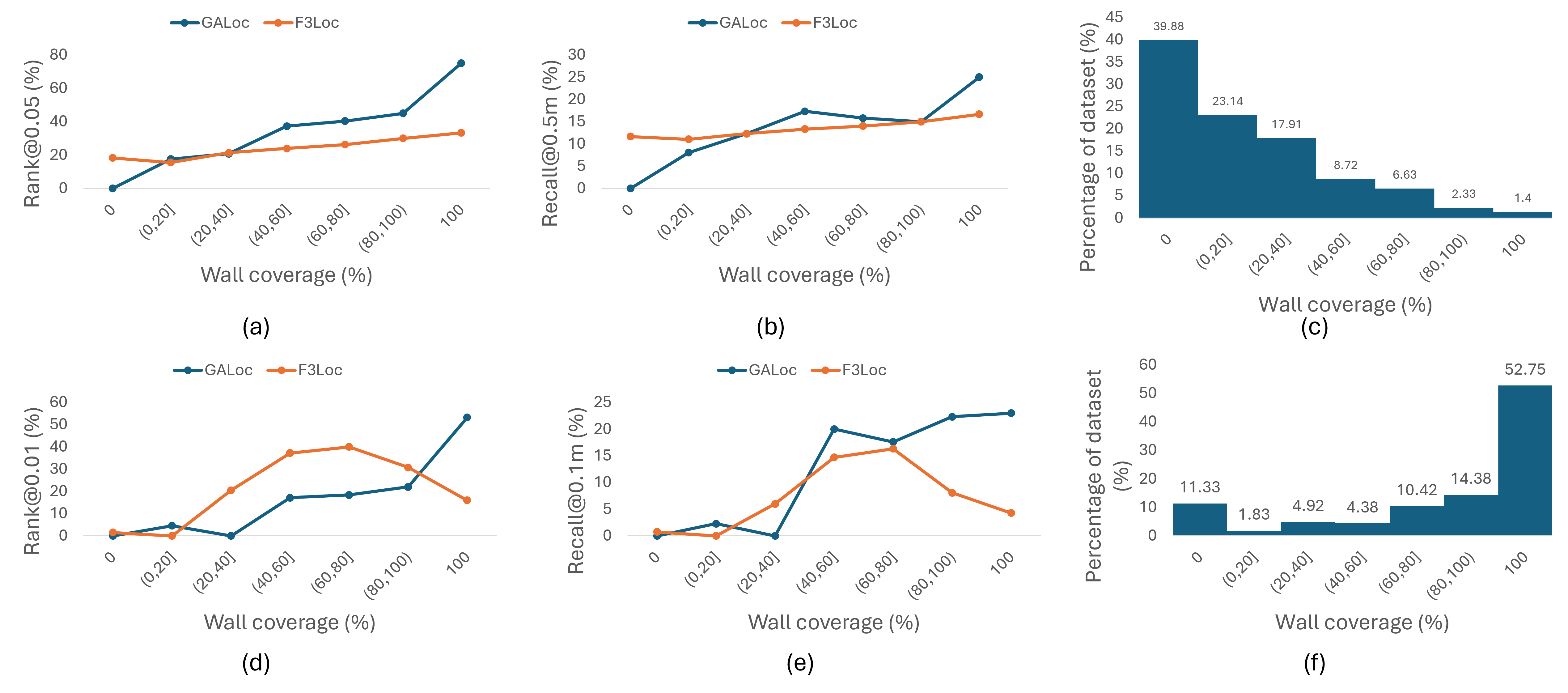}
    \caption{\textbf{Coverage-conditioned analysis comparing $\text{GALoc}_{(\sigma=5)}$ and $\text{F3Loc}_m$.}
    Wall coverage is the fraction of horizontal FOV spanned by fully reconstructed walls ($K$).
    \emph{Top:} Structured3D, \emph{Bottom:} Gibson.
    \textbf{(a,d)}~Rank at 0.05\% and 0.01\%. \textbf{(b,e)}~Recall at 0.5\,m and 0.1\,m. \textbf{(c,f)}~Dataset distribution.
    Nearly 40\% of Structured3D frames have zero wall coverage, explaining GALoc's aggregate gap (\Cref{tab:combined_localization_unified}); on Gibson, over 80\% of frames fall in the ${>}40\%$ coverage bins where GALoc leads on recall (e,f).}
    \label{fig:coverage_combined}
    \vspace{-20pt}
\end{figure}

\noindent \Cref{tab:combined_localization_unified} reports single-image results on S3D~\cite{zheng2020structured3d} and Gibson~\cite{xia2018gibson}. 
\noindent $F3Loc_m$ (the single-image variant of F3Loc~\cite{chen2024f3loc}; $F3Loc_{comp}$ denotes its multi-frame complete variant used in the sequential comparison) reports higher aggregate recall on S3D because it produces an estimate for every frame, including structure-blind close-ups where no wall geometry exists.
GALoc instead reports zero confidence there (\Cref{fig:single_image_localization}). 
This is a trade-off rather than a strict advantage. 
The gap is concentrated in low-coverage frames: nearly 40\% of Structured3D frames have zero wall coverage, and where coverage exceeds ${\sim}40\%$ GALoc matches or surpasses $F3Loc_m$ at the thresholds shown in \Cref{fig:coverage_combined}. 
Gibson presents the complementary distribution---over $80\%$ of frames have sufficient coverage, reflecting both its wider FOV ($\sim\!108^\circ$ vs $\sim\!80^\circ$) and the topological completeness of our junctions.
GALoc leads accordingly, achieving $18.1\%$ recall at $0.1$\,m ($3\times$ $F3Loc_m$) and $45.3\%$ rank at $0.01\%$ versus $19.8\%$ in aggregate.
 \subsection{Sequential Localization}

\label{subsec:sequential_loc}
\begin{figure*}[t]
    \centering
    \includegraphics[width=0.8\textwidth]{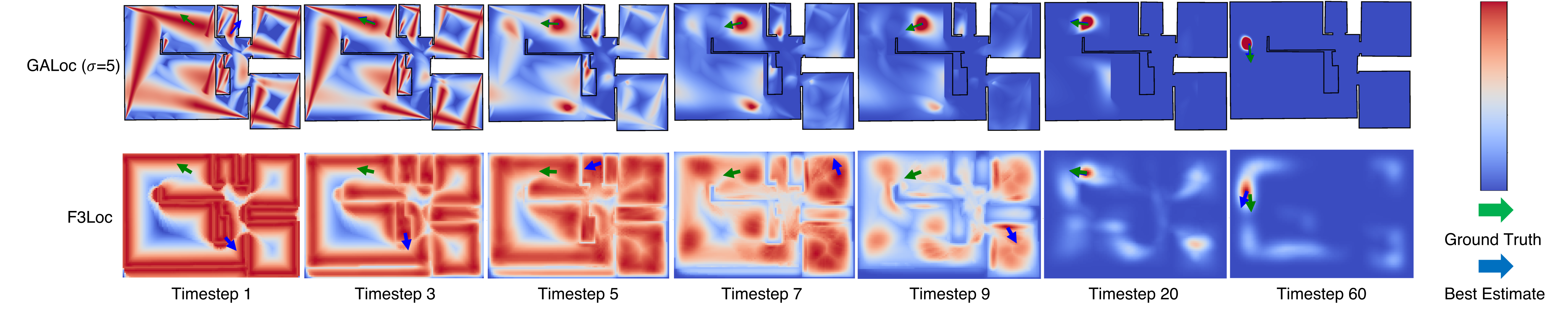}
    \caption{\textbf{Sequential localization on Gibson.}
    $GALoc_{(\sigma=5)}$ converges to the ground-truth pose within 7--10 frames; F3Loc~\cite{chen2024f3loc} typically requires 20--50 frames and exhibits less stable convergence.}
    \label{fig:cumulative_image_localization}
    \vspace{-13pt}
\end{figure*}

\begin{figure}
    \centering
    \includegraphics[width=0.8\columnwidth]{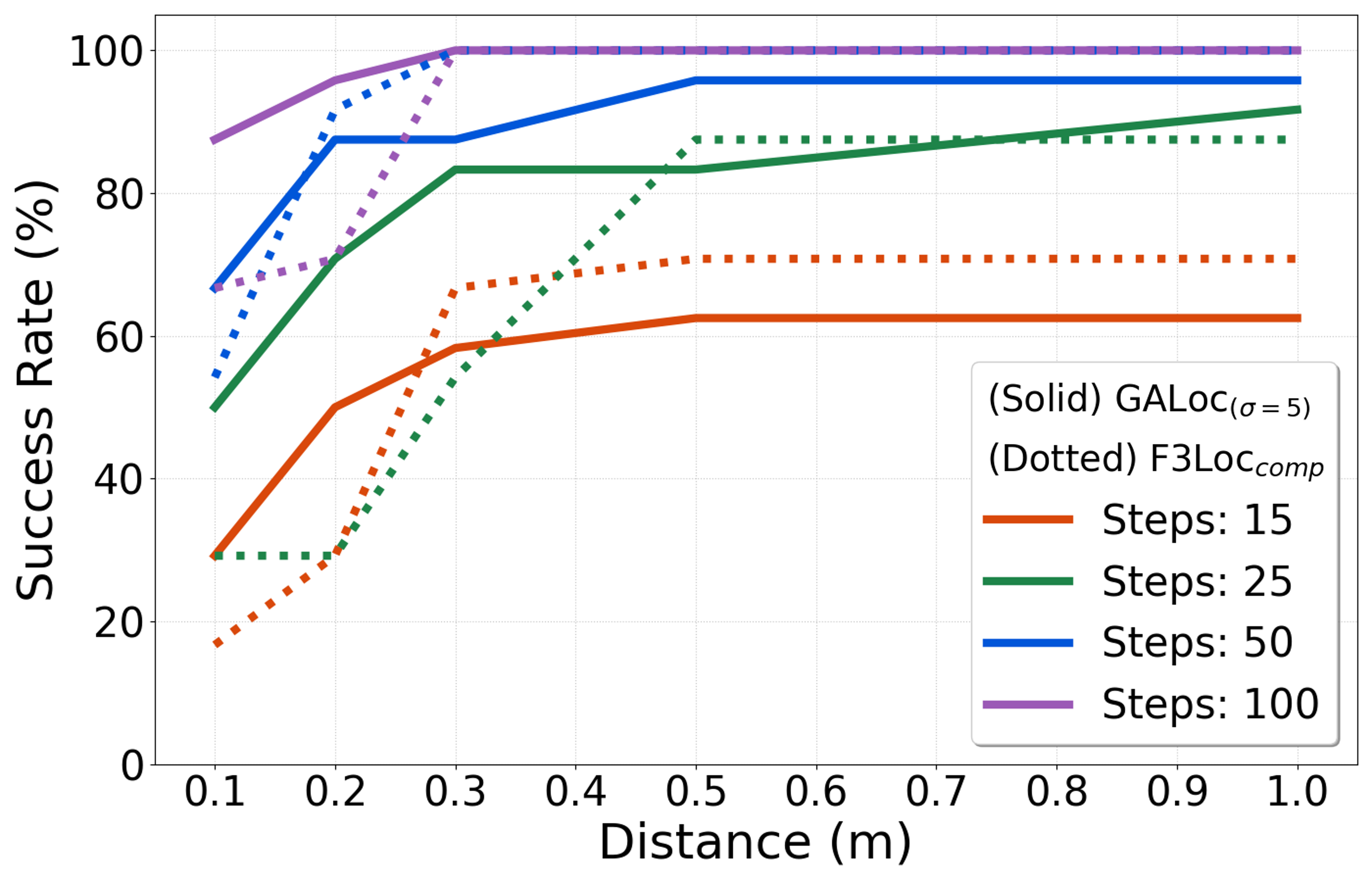}
    \caption{\textbf{Sequential localization on Gibson.} Success rate vs.\ distance threshold for various sequence lengths. Solid: $GALoc_{(\sigma=5)}$. Dotted: $F3Loc_{comp}$. At short horizons GALoc trails slightly as few frames lead to geometric ambiguities; once evidence accumulates, GALoc overtakes F3Loc and holds 88\% vs.\ 68\% SR at 0.1\,m with 100 steps.}
    \label{fig:cumulative_localization_results}
    \vspace{-15pt}
\end{figure}

\noindent Temporal fusion resolves single-frame ambiguities, making this the most realistic deployment setting.
We compare against F3Loc only: PALMS+~\cite{cheng2025palms+} produces a single heatmap from a stationary scan and tracks via odometry alone, and SemRayLoc~\cite{grader2025supercharging} requires semantic annotations Gibson lacks.
We evaluate sequence lengths of 15, 25, 50, and 100 on Gibson; S3D provides no sequential data.

\noindent Both $F3Loc_{comp}$ and $GALoc_{\sigma=5}$ improve with sequence length (\Cref{fig:cumulative_image_localization,fig:cumulative_localization_results}); the distinction is convergence speed and precision.
F3Loc holds a slight edge at 15 steps, where few observations are available, but loses it rapidly: its low-coverage wall-distance estimates inject confident-but-wrong observations that corrupt the filter.
GALoc instead abstains when no panel is fully recovered, so high-coverage frames accumulate geometric evidence while zero-coverage ones propagate uncertainty without misinforming the belief.
\Cref{fig:single_image_localization}~(bottom) shows this directly: too close to a wall, GALoc's heatmap is uniform, whereas F3Loc's is diffuse but spatially biased, steering the belief away from the true pose.


\subsection{Real-World Evaluation}
\label{sec:real_world}

\begin{table}[htbp]
  \scriptsize
  \centering
  \caption{Real-world sequential localization with GALoc and F3Loc (both zero-shot) across five sequences in two environments (three in office, two in corridor).}
  \label{tab:real_world_sequential}
  \vspace{-5pt}
  \begin{tabular}{l cccccc}
    & \multicolumn{6}{c}{\textbf{Recall (distance thresholds)}} \\
    \cmidrule(lr){2-7}
    \textbf{Method} & 0.1m & 0.3m & 0.5m & 0.7m & 1m, $30^\circ$ & 1m \\
    \midrule
    GALoc  & 0.0 & 0.0 & 0.0 & 20.0 & 20.0 & 20.0 \\
    F3Loc  & 0.0 & 0.0 & 0.0 &  0.0 &  0.0 &  0.0 \\
  \end{tabular}
  \vspace{-20pt}
\end{table}

\begin{figure*}[t]
    \centering
    \includegraphics[width=0.8\textwidth]{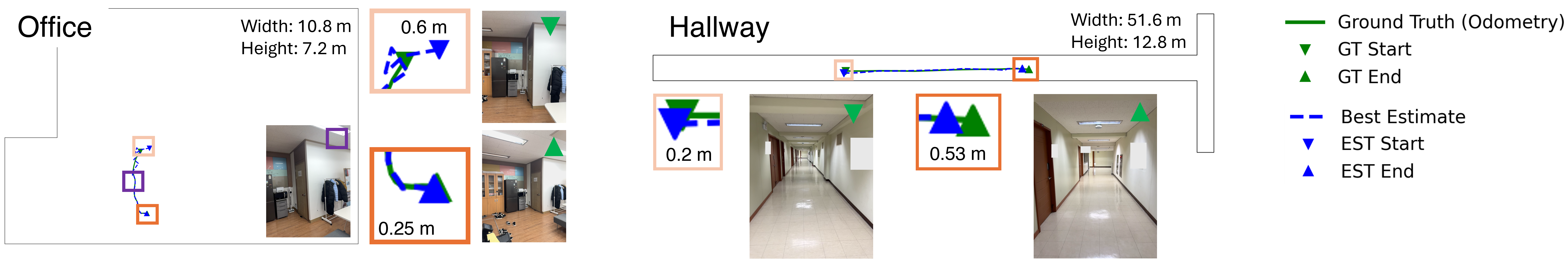}
    \caption{\textbf{Real-world sequential localization.} Trajectories in an office ($10.8\!\times\!7.2$\,m, top) and a long corridor ($51.6\!\times\!12.8$\,m, bottom), with endpoint errors of 0.25--0.6\,m and 0.2--0.53\,m respectively. Photos show viewpoints along each sequence.}
    \label{fig:real_world_sequential_results}
    \vspace{-10pt}
\end{figure*}

\begin{figure*}[t]
    \centering
    \includegraphics[width=0.8\linewidth]{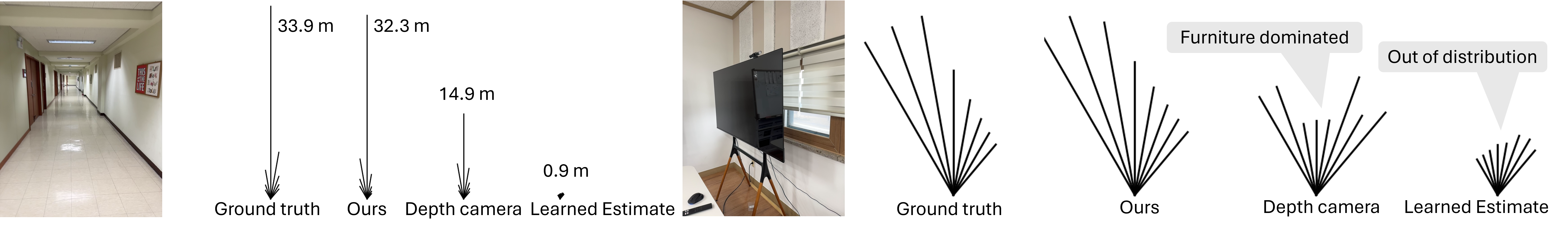}
    \caption{\textbf{Real-world ray comparison.} Top: office. Bottom: corridor. GALoc recovers wall structure directly, with rays reaching about 32 m in the corridor. Metric scale (from ceiling height) is for visualization only; matching is scale-invariant.}
    \label{fig:real_world_rays}
    \vspace{-20pt}
\end{figure*}

\noindent We evaluate GALoc on five sequences collected in two environments---a cluttered office ($10.8\!\times\!7.2$\,m) and a long corridor ($51.6\!\times\!12.8$\,m)---using an iPhone with ARKit providing visual-inertial odometry as semi-ground-truth. \Cref{tab:real_world_sequential} reports the zero-shot results with the automatic wireframe detector: GALoc reaches 20\% recall at 1\,m, while F3Loc's recall is zero at all thresholds---its learned wall-distance model, despite being trained to see through clutter, fails entirely on these out-of-distribution environments (see \Cref{fig:real_world_rays}). 
With hand-labeled junctions (semi-GT), GALoc instead reaches 80\% recall at 1\,m on the same sequences, so the gap between the two settings isolates detector domain transfer, not the localization geometry, as the bottleneck.

\noindent \Cref{fig:real_world_sequential_results} shows two sequences, with endpoint errors of 0.25--0.6\,m (office) and 0.2--0.53\,m (corridor).
\Cref{fig:real_world_rays} shows why depth alternatives fail: GALoc's rays reach 32--34\,m in corridors, whereas F3Loc's estimate collapses on these out-of-distribution frames and, in offices, is dominated by furniture.
Consumer RGB-D sensors also saturate near ${\sim}$15\,m, matching F3Loc's~\cite{chen2024f3loc} maximum range.

 \subsection{Runtime Comparison}

\label{subsec:runtime}

\begin{table}[h]
    \vspace{-10pt}
    \caption{Runtime (i5-12400F, 32GB RAM, RTX3060).}
    \scriptsize
    \centering
    \setlength{\tabcolsep}{3pt} 
    \begin{tabular}{lcccc}
    Method & Wireframe/Depth & Gravity align & Matching & End-to-end \\
    \midrule
    GALoc & 2.10\,s & 0.01\,s & 0.19\,s & 2.32\,s \\
    F3Loc & 0.04\,s & 0.01\,s & 0.34\,s & 0.40\,s \\
    \end{tabular}
    \label{tab:runtime}
    \vspace{-15pt}
\end{table}

\noindent \Cref{tab:runtime} reports runtime.
GALoc uses Numba-based parallelization and evaluates a dense $0.01\,\text{m},\,1^{\circ}$ grid.

\subsection{Ablation: Robustness to Input Noise}
\label{sec:ablation}

\noindent Perturbing each junction independently makes $A(\theta)$ full-rank at every $\theta$, so the gauge is recovered from a system no angle satisfies exactly.
Degrading the perturbation from $\sigma{=}3$ (96.76\%) to $\sigma{=}7$ (92.78\%) costs 4.2\,pp in recall at 1\,m (\Cref{tab:combined_localization_unified}, right): noise raises $\sigma_{\min}(A(\hat\theta))$ and shifts $\hat\theta$, and the EYM projection still returns an architecturally valid wireframe at every noise level, so the BEV layout degrades rather than collapsing.
\section{Limitations and Conclusion} 


\noindent\textbf{Limitations.}
GALoc assumes Atlanta-world geometry~\cite{schindler2004atlanta} and abstains when wall structure is not visible, reducing single-frame coverage compared to depth-based methods (\Cref{fig:single_image_localization}).
Its accuracy is bounded by the wireframe extractor, which also dominates runtime.
Our real-world evaluation is also limited in scope.
Two representative failure modes are zero wall coverage (camera too close to a wall, all FOV arcs unknown, filter falls back to odometry) and partial junction detection ($j_\mathrm{true}\le 1$, panel discarded, reduced known-FOV arc).
The second is the harder case: unlike zero coverage, where GALoc abstains, the surviving panels still yield a confident but geometrically incomplete BEV layout.
Fusing wireframe geometry with depth---depth for scale and close range, wireframes for long-range structural constraints---is a natural extension, as is comparison against uncertainty-aware depth pipelines under a common front-end.

\noindent\textbf{Conclusion.}
We presented GALoc, a depth-free floorplan localization framework that recovers gravity-aligned wireframes via null-space search and matches them against the floorplan through metric-free SE(2) search.
GALoc and depth-based methods occupy complementary regimes: when wall geometry is visible, GALoc achieves $3{\times}$ the single-image recall at 0.1\,m and 88\% vs.\ 68\% sequential success on Gibson, while abstaining when it is not.
GALoc's accuracy is bounded by the wireframe front-end: our Gibson protocol supplies junctions with correct topology and ceiling/floor labels, which current detectors do not always meet---the zero-shot vs hand-labeled real-world gap ($20\%$ vs $80\%$ recall at $1$\,m) shows the cost.
As wireframe extraction matures toward the robustness monocular depth enjoys today, geometry-first localization stands to become competitive across a wider range of environments.
\vspace{-5pt}
\bibliographystyle{IEEEtran}
\bibliography{jeahnhan_ral2025}



\end{document}